\pdfoutput=1

\documentclass[11pt]{article}
\usepackage[table]{xcolor}
\usepackage{hyperref}

\usepackage[final]{acl}

\usepackage{times}
\usepackage{latexsym}
\usepackage{multirow}
\usepackage{amsmath}
\usepackage{authblk}
\usepackage{graphicx}

\usepackage[T1]{fontenc}

\usepackage[utf8]{inputenc}

\usepackage{microtype}

\usepackage{inconsolata}

\usepackage{graphicx}

\definecolor{darkgreen}{rgb}{0.0, 0.5, 0.0}
\definecolor{darkred}{rgb}{0.5, 0.0, 0.0}

\usepackage{tikz}
\usetikzlibrary{shapes.geometric, arrows, positioning}

\tikzstyle{startstop} = [rectangle, rounded corners, minimum width=3cm, minimum height=1cm,text centered, draw=black, fill=red!30]
\tikzstyle{process} = [rectangle, minimum width=3cm, minimum height=1cm, text centered, draw=black, fill=orange!30]
\tikzstyle{decision} = [diamond, minimum width=3cm, minimum height=1cm, text centered, draw=black, fill=yellow!30]
\tikzstyle{arrow} = [thick,->,>=stealth]

\title{Fast Prompt Alignment for Text-to-Image Generation}

\author[1]{Khalil Mrini}
\author[1]{Hanlin Lu}
\author[1]{Linjie Yang}
\author[1]{Weilin Huang}
\author[1]{Heng Wang}

\affil[1]{TikTok\\
  \texttt{\{khalil.mrini, hanlin.lu, linjie.yang, xuhu.ai\}@tiktok.com, hengwang00@gmail.com}}

\begin{document}
\maketitle

\begin{abstract}
Text-to-image generation has advanced rapidly, yet aligning complex textual prompts with generated visuals remains challenging, especially with intricate object relationships and fine-grained details. This paper introduces \textbf{Fast Prompt Alignment (FPA)}, a prompt optimization framework that leverages a one-pass approach, enhancing text-to-image alignment efficiency without the iterative overhead typical of current methods like OPT2I. FPA uses large language models (LLMs) for single-iteration prompt paraphrasing, followed by fine-tuning or in-context learning with optimized prompts to enable real-time inference, reducing computational demands while preserving alignment fidelity. Extensive evaluations on the COCO Captions and PartiPrompts datasets demonstrate that FPA achieves competitive text-image alignment scores at a fraction of the processing time, as validated through both automated metrics (TIFA, VQA) and human evaluation. A human study with expert annotators further reveals a strong correlation between human alignment judgments and automated scores, underscoring the robustness of FPA’s improvements. The proposed method showcases a scalable, efficient alternative to iterative prompt optimization, enabling broader applicability in real-time, high-demand settings. The codebase is provided to facilitate further research.\footnote{Github link: \url{https://github.com/tiktok/fast_prompt_alignment}}
\end{abstract}

\section{Introduction}

Text-to-image generation models, such as Stable Diffusion and DALL-E, have made significant advances in generating high-quality images from textual descriptions. Despite these advancements, models often struggle to maintain consistent alignment between complex prompts and the resulting images, particularly when dealing with intricate object relationships or fine-grained details. Prompt optimization techniques have been introduced to address these challenges, with the goal of refining input prompts to improve alignment between text and generated images. 

\begin{figure}[h!]
    \centering
    \includegraphics[width=\columnwidth]{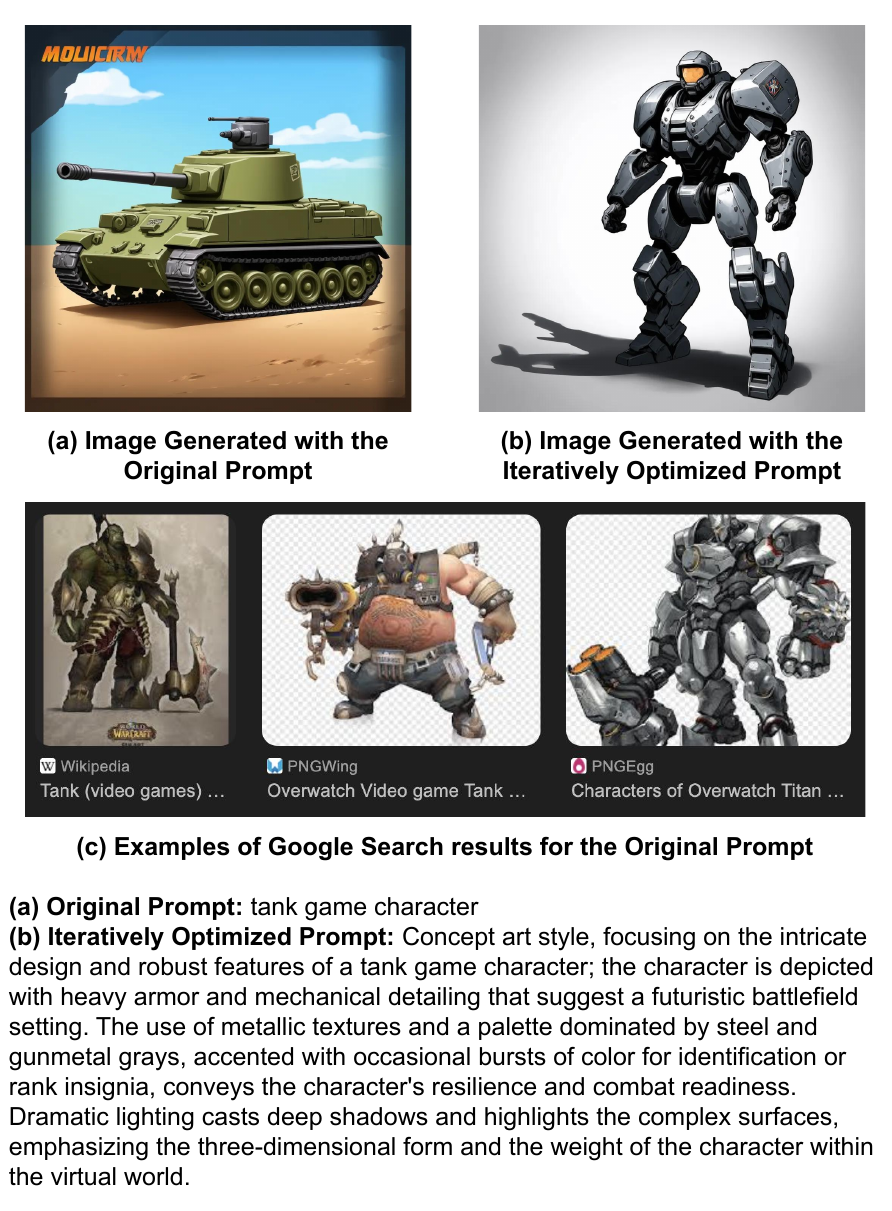}
    \caption{Example of iterative optimization for a prompt used to generate images. The model we use is Stable Diffusion 3 Medium. We provide Google image search results for the original prompt for comparison.}
    \label{fig:example_fpa}
\end{figure}

One prominent approach to prompt optimization is \textit{OPT2I} \cite{manas2024opt2i}, which iteratively generates paraphrases of an input prompt, evaluates the corresponding generated images using metrics like TIFA \cite{hu2023tifa} and CLIP \cite{radford2021learning}, and selects the paraphrase that produces the best results. While \textit{OPT2I} achieves notable improvements in image alignment, its iterative nature makes it computationally expensive and slow. Each prompt requires multiple rounds of paraphrasing and evaluation, making the method less feasible for real-time or large-scale applications.

To address these limitations, we propose \textit{Fast Prompt Alignment (FPA)}, a method that accelerates the process of prompt optimization without sacrificing alignment quality. Instead of relying on multiple iterations, FPA takes the results from iterative optimization processes like \textit{OPT2I} and uses them to train a model capable of real-time prompt optimization. This approach allows FPA to offer a computationally efficient alternative, providing near-instant alignment improvements with performance close to, but faster than, full iterative methods.

Our method leverages large language models (LLMs) such as GPT-4o\footnote{For clarification, we use the Azure-provided API for OpenAI's GPT-4o. The ubiquitous UI version is ChatGPT-4o.} for prompt paraphrasing, where a single iteration of paraphrasing is sufficient. We then fine-tune smaller LLMs (e.g., 7B parameters) on the optimized prompts derived from the iterative process. This fine-tuning enables real-time inference with minimal overhead while still producing high-quality results. We find that the alignment achieved by FPA on a 7B-parameter LLM is lower, even when fine-tuned with iteratively optimized prompts, than a fully iterative approach like \textit{OPT2I}. However, FPA on a 123B-parameter LLM with in-context learning demonstrates competitive performance while offering significant speed improvements \cite{rombach2022high}.

In our experiments, we find that FPA significantly reduces the time required for prompt optimization. While it seems that we are unable to transfer the learnings from iterative prompt optimization to a 7B-parameter LLM, simple in-context learning with 100 examples on a 123B-parameter LLM provides high scores in TIFA and VQA scores, well above the scores obtained by the user-generated original prompts. Additionally, we conducted a human evaluation with expert annotators, who assessed text-to-image alignment and image structure across optimized and original prompts. The results showed a strong positive correlation between human judgment and automated metrics (TIFA and VQA), validating the alignment improvements achieved by FPA. By significantly reducing the time required for prompt optimization and demonstrating competitive performance, FPA offers a scalable, efficient alternative suitable for real-time, high-demand settings.

\section{Related Work}

Prompt optimization has become a critical research area in text-to-image generation, aiming to enhance the alignment between complex textual inputs and the generated images. Several methods have been proposed to improve this alignment, with varying approaches in terms of paraphrasing, reinforcement learning (RL), and fine-tuning large language models (LLMs).

One of the prominent methods is \textit{Promptist} \cite{hao2023promptist}, which uses reinforcement learning combined with a language model to iteratively rephrase prompts. By generating more detailed and informative prompts, \textit{Promptist} improves the text-to-image consistency across different datasets, including DiffusionDB, MS COCO \cite{chen2015microsoft}, and ImageNet-21k \cite{deng2009imagenet}. This method fine-tunes a GPT model using supervised data and leverages RL to explore better paraphrases, demonstrating significant gains in image fidelity.

Another approach, \textit{Dynamic Prompt Optimizing}, explores multimodal prompting and refinement to generate better text-image alignments \cite{mo2024promptcharm}. This method provides interactive ways for users to refine their prompts in real-time, enhancing the control over image generation through multiple iterations of prompt refinement.

Gradient-based optimization methods have also been applied to prompt tuning, as demonstrated in \textit{Hard Prompts Made Easy} \cite{wen2023hardprompts}. This method enables the optimization of hard prompts for text-to-image models without requiring extensive manual tuning. By using a gradient-based approach to search for optimal prompts, this method bypasses traditional token-level content filters and improves alignment across datasets such as LAION \cite{schuhmann2022laion} and Celeb-A \cite{liu2015faceattributes}.

The idea of improving prompt alignment through few-shot learning was proposed by \cite{yang2023fewshotprompts}, where in-context learning is used to guide the model in generating better prompts. This method utilizes prompt templates and a few example prompts to significantly improve the fidelity and aesthetics of the generated images. The use of GPT-J in combination with a few-shot learning paradigm allows the model to generate prompts that align better with the text-to-image generation tasks.

Another key method, \textit{NegOpt} \cite{zhou2024negoptprompts}, focuses on optimizing negative prompts to remove undesirable characteristics from generated images. This method uses supervised fine-tuning and reinforcement learning to optimize these negative prompts, significantly improving the visual quality and coherence of the generated images.

Our work is closely related to \textit{OPT2I} \cite{manas2024opt2i}, which iteratively optimizes prompts using an LLM and automatic scoring metrics such as TIFA \cite{hu2023tifa} and CLIP \cite{radford2021learning}. While effective in improving text-image consistency, \textit{OPT2I} suffers from computational inefficiencies due to its iterative nature. In contrast, our proposed \textit{Fast Prompt Alignment (FPA)} reduces the time and computational cost by leveraging results from iterative optimization to train a fine-tuned model capable of real-time inference, and to use in-context learning with a large LLM. This last approach achieves performance comparable to \textit{OPT2I} but with significantly reduced computational overhead.

\section{Methodology}

Fast Prompt Alignment (FPA) is a prompt optimization method designed to enhance the efficiency of text-to-image generation models. Leveraging the results of iterative methods such as OPT2I, which require multiple rounds of paraphrasing and evaluation, FPA optimizes the process with a single inference pass. This section outlines the key steps of FPA: paraphrase generation, image generation, scoring, fine-tuning and inference for a smaller LLM (7B parameters), or just inference with in-context learning for a larger LLM (123B parameters).

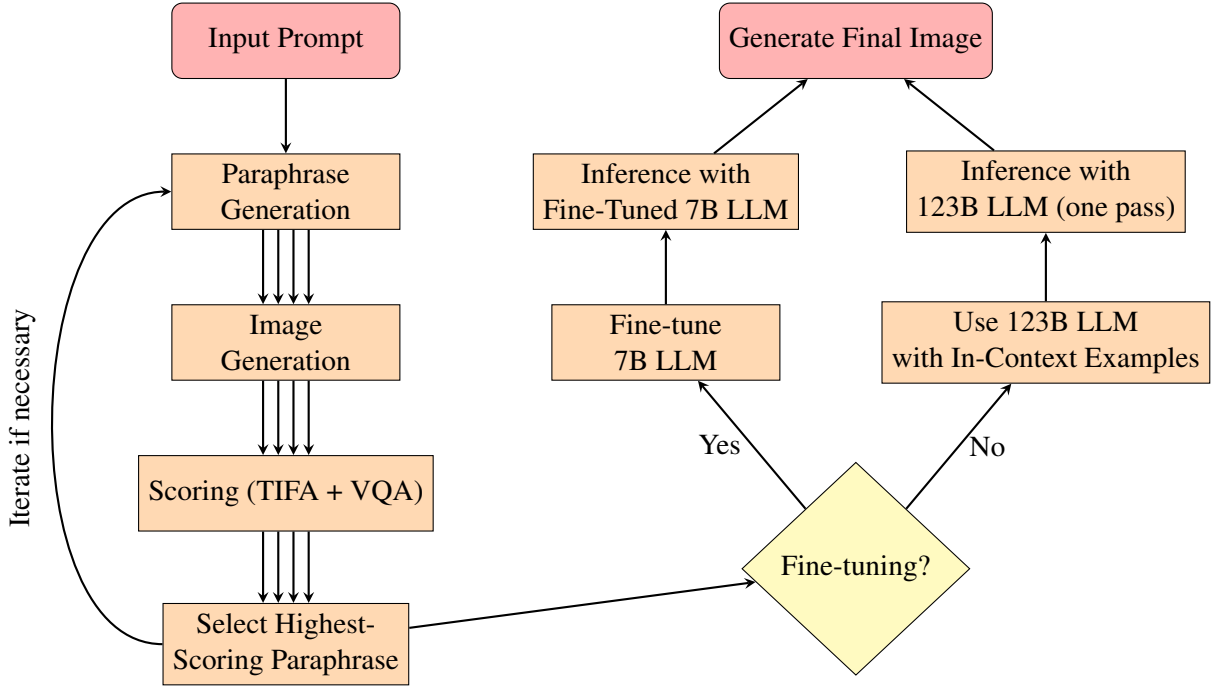
\begin{figure*}[h!]
    \centering
    \begin{tikzpicture}[node distance=2cm]

        \tikzstyle{startstop} = [rectangle, rounded corners, minimum width=3cm, minimum height=1cm, text centered, draw=black, fill=red!30]
        \tikzstyle{process} = [rectangle, minimum width=3cm, minimum height=1cm, text centered, draw=black, fill=orange!30, align=center]
        \tikzstyle{decision} = [diamond, minimum width=3cm, minimum height=1cm, text centered, draw=black, fill=yellow!30]
        \tikzstyle{arrow} = [thick,->,>=stealth]

        \node (start) [startstop] {Input Prompt};
        \node (generate) [process, below of=start] {Paraphrase\\ Generation};
        \node (imagegen) [process, below of=generate] {Image\\ Generation};
        \node (scoring) [process, below of=imagegen] {Scoring (TIFA + VQA)};
        \node (select) [process, below of=scoring] {Select Highest-\\Scoring Paraphrase};

        \node (finetune_process) [process, right of=imagegen, xshift=3cm] {Fine-tune\\ 7B LLM};
        \node (finetune) [decision, right of=scoring, yshift=-1cm, xshift=5.5cm] {Fine-tuning?};
        \node (inference_small) [process, right of=generate,  xshift=3cm] {Inference with\\ Fine-Tuned 7B LLM};

        \node (incontext) [process, right of=finetune_process, xshift=3cm] {Use 123B LLM\\ with In-Context Examples};
        \node (inference_large) [process, right of=inference_small, xshift=3cm] {Inference with\\ 123B LLM (one pass)};

        \node (final_image) [startstop, right of=start, xshift=5.5cm] {Generate Final Image};

        \draw [arrow] (start) -- (generate);

        \foreach \i in {0.3cm, 0.1cm, -0.1cm, -0.3cm} {
            \draw [arrow] ([xshift=\i] generate.south) -- ([xshift=\i] imagegen.north);
            \draw [arrow] ([xshift=\i] imagegen.south) -- ([xshift=\i] scoring.north);
            \draw [arrow] ([xshift=\i] scoring.south) -- ([xshift=\i] select.north);
        }

        \draw [arrow] (select) -- (finetune);
        \draw [arrow] (finetune) -- node[anchor=east] {Yes} (finetune_process);
        \draw [arrow] (finetune_process) -- (inference_small);

        \draw [arrow] (finetune) -- node[anchor=west] {No} (incontext);
        \draw [arrow] (incontext) -- (inference_large);

        \draw [arrow] (inference_small) -- (final_image);
        \draw [arrow] (inference_large) -- (final_image);

        \draw [arrow] (select.west) .. controls +(left:2cm) and +(left:2cm) .. node[rotate=90, anchor=center, yshift=0.4cm] {Iterate if necessary} (generate.west);

    \end{tikzpicture}
    \caption{Diagram of the Fast Prompt Alignment (FPA) method for text-to-image generation. The process includes paraphrase generation, image scoring, and both fine-tuning and in-context learning methods for efficient prompt optimization. We generate 4 paraphrases in our case, and we illustrate that with the 4 parallel arrows.}
    \label{fig:fpa_method}
\end{figure*}

\subsection{Overview of FPA Architecture}

The FPA architecture follows these steps:
\begin{itemize}
    \item \textbf{Paraphrase Generation}: A large language model (LLM) generates several paraphrases of the input prompt.
    \item \textbf{Image Generation}: Each paraphrase is used by a frozen text-to-image model to generate images.
    \item \textbf{Scoring}: The generated images are evaluated using automated metrics, and the highest-scoring paraphrase is selected.
    \item \textbf{Fine-tuning and Inference on a smaller LLM}: The selected paraphrases are used to fine-tune a smaller LLM to internalize the optimization process. The fine-tuned model performs real-time prompt optimization without requiring iterative refinements.
    \item \textbf{Inference with In-Context Learning on a larger LLM}: This method enables prompt optimization in one pass by leveraging a random set of example pairs, reducing the need for iterative refinement and enhancing computational efficiency. The example pairs consist of user-generated prompts and their iteratively optimized counterparts.
\end{itemize}

Steps 1, 2, and 3 may be repeated using the highest-scoring paraphrase from the previous round. This architecture is shown in Figure~\ref{fig:fpa_method}.

\subsection{Paraphrase Generation}

Given an initial prompt \(P_0\), a large LLM (e.g., GPT-4o) generates multiple paraphrases \( \mathcal{P} = \{P_1, P_2, \dots, P_n\} \), each of which expresses the meaning of the input prompt in a different way. The goal of this step is to explore alternative formulations of the prompt, which may lead to improved text-to-image alignment. These paraphrases are generated in a single iteration, with the option to repeat the process using the highest-scoring paraphrase to further improve alignment.

\subsection{Image Generation and Scoring}

Each paraphrase \(P_i \in \mathcal{P}\) is used by the frozen text-to-image model \(M\) (e.g., Stable Diffusion 3.0) to generate an image \(I_i\) as follows:
\[
I_i = M(P_i)
\]
The generated images are scored based on their alignment with the prompts using two key metrics:
\begin{itemize}
    \item \textbf{TIFA (Text-to-Image Faithfulness)}: The TIFA score \cite{hu2023tifa} is computed by generating question-answer pairs from the input text prompt using a language model, in our case GPT-4o, which targets specific elements of the image like objects, attributes, and relationships. These questions are then answered by a visual-question-answering model that analyzes the image content. The final TIFA score represents the accuracy of the VQA model's answers when compared to the expected answers based on the text prompt, providing a measure of how faithfully the image reflects the input text, with a score ranging from 0 to 1.
    \item \textbf{VQA (Visual Question Answering)}: VQAScore \cite{lin2024evaluating} is computed by converting the input text prompt into a question format, such as: ``Does this figure show \{text\}?'', where the \textit{text} corresponds to one of the noun chunks mentioned in the prompt, as detected by a call to an LLM. The image and the corresponding question are then passed through a VQA model, which predicts the probability that the answer is ``Yes'' using an answer decoder. The final VQAScore is the probability that the VQA model will generate a ``Yes'' answer, averaged for all noun chunks, serving as an alignment metric between the input text and the generated image. In this case again, we use GPT-4o and get the probability from the OpenAI API.
\end{itemize}

The final score for the $i$-th paraphrase \(P_i\) and image \(I\) is:
\[
S(P_i, I) = \text{TIFA}(P_i, I) + \text{VQA}(P_i, I)
\]
The paraphrase with the highest score is selected, and steps 2 and 3 can be repeated with the highest-scoring paraphrase from the previous round.

\subsection{Fine-tuning and Inference with a 7B-parameter LLM}

After identifying the best paraphrase \(P^*\), FPA fine-tunes a smaller LLM (7B parameters) using the top-performing paraphrases. This fine-tuning step enables the smaller LLM to internalize the optimization process, allowing it to generate optimized prompts during inference. The fine-tuning objective aims to minimize the probability distribution difference between the paraphrases generated by the smaller LLM and the best-scoring paraphrases from the larger LLM, and is defined as:
\[
\mathcal{L}_{\text{SFT}} = -\sum_{t=1}^{m} \log P_\theta(y_t | y_{<t}, \mathbf{x})
\]
where \((y_1, ..., y_m)\) is the target output, corresponding to the best-scoring paraphrase.

The fine-tuned smaller LLM is then used for real-time inference. For a new prompt \(P_{\text{new}}\), the fine-tuned model generates an optimized paraphrase \(P_{\text{opt}}\) in one pass:
\[
P_{\text{opt}}^{\text{one-pass}} = \text{LLM}_{\text{small}}^{\text{fine-tuned}}(P_{\text{new}})
\]

The optimized paraphrase is passed to the text-to-image model to generate the final image:
\[
I_{\text{opt}} = M(P_{\text{opt}}^{\text{one-pass}})
\]

\subsection{Inference with In-Context Learning on a 123B-parameter LLM}

In lieu of fine-tuning, we can leverage in-context learning by using a random set of \( n \) example pairs \((P_{\text{orig}}, P_{\text{opt}})\), where \( P_{\text{orig}} \) is a user-generated original prompt, and \( P_{\text{opt}} \) is the corresponding optimized prompt obtained through iterative optimization. By conditioning a 123B-parameter LLM with these \( n \) example pairs, the model can generalize the optimization pattern and produce an optimized prompt \( P_{\text{opt}}^{\text{one-pass}} \) directly from a new input prompt \( P_{\text{new}} \), bypassing the need for iterative refinement.

Define:
\begin{itemize}
    \item The example set as \( \{(P_{\text{orig}}^{(i)}, P_{\text{opt}}^{(i)})\}_{i=1}^{n} \)
    \item The optimization function learned by the LLM as \( f_{\text{LLM}} \)
\end{itemize}

Then, given a new prompt \( P_{\text{new}} \), the optimized prompt \( P_{\text{opt}}^{\text{one-pass}} \) can be generated in one pass as:

\[
P_{\text{opt}}^{\text{one-pass}} = f_{\text{LLM}}(P_{\text{new}}; \{(P_{\text{orig}}^{(i)}, P_{\text{opt}}^{(i)})\}_{i=1}^{n})
\]

This single-pass optimization reduces computational complexity and leverages in-context examples for real-time prompt enhancement, thus achieving efficient and scalable prompt optimization suitable for large-scale applications.

\subsection{Computational Efficiency}

One of the primary advantages of FPA is its computational efficiency. Iterative methods like OPT2I typically require multiple rounds of paraphrasing and scoring, with a complexity of \(O(n \cdot m)\), where \(n\) is the number of iterations and \(m\) is the number of paraphrases per iteration. In contrast, FPA reduces the complexity to \(O(m)\) for paraphrasing and \(O(1)\) for inference, making it suitable for real-time applications.

\section{Experiments}

\begin{table*}[h!]
    \centering
    \begin{tabular}{|l|l|c|c|c|}
        \hline
        \textbf{Dataset} & \textbf{Mode} & \textbf{TIFA Score} & \textbf{VQA Score} & \textbf{Average} \\
        \hline
        \multirow{6}{*}{\textbf{Coco Captions}} 
            & Original Prompts & 0.863 & 0.829 & 0.846 \\
            & After 2 Iterations & 0.967 & 0.954 & 0.961 \\ \cline{2-5}
            & Mistral 7B & 0.845 & 0.850 & 0.848 \\
            & Fine-tuned Mistral 7B & 0.839 & 0.811 & 0.825 \\ \cline{2-5}
            & Mistral Large (no examples) & 0.882 & 0.873 & 0.878 \\
            & Mistral Large (100 examples) & 0.872 & 0.867 & 0.869 \\
        \hline
        \multirow{6}{*}{\textbf{PartiPrompts}} 
            & Original Prompts & 0.813 & 0.760 & 0.787 \\
            & After 2 Iterations & 0.937 & 0.930 & 0.934 \\ \cline{2-5}
            & Mistral 7B & 0.799 & 0.761 & 0.780 \\
            & Fine-tuned Mistral 7B & 0.820 & 0.784 & 0.802 \\ \cline{2-5}
            & Mistral Large (no examples) & 0.844 & 0.812 & 0.828 \\
            & Mistral Large (100 examples) & 0.856 & 0.833 & 0.845 \\
        \hline
    \end{tabular}
    \caption{Comparison of TIFA and VQA scores for CocoCaptions and PartiPrompt datasets under different modes: original prompts versus after two iterations of iterative prompt optimization. The combined score is the average of the TIFA and VQA scores.}
    \label{tab:dataset_optimization_results}
\end{table*}

\subsection{Datasets}

For our experiments, we utilize three datasets with distinct characteristics—COCO Captions, PartiPrompts, and MidJourney Prompts—to evaluate the performance of FPA. We use the MidJourney Prompts for training, while COCO Captions and PartiPrompts are reserved for testing.

\paragraph{COCO Captions:} The COCO Captions dataset \cite{chen2015microsoft} consists of over 330,000 images, each accompanied by five human-generated captions. For our experiments, we use the well-known 2,000-image subset defined by \citet{karpathy2015deep}, which has become a standard benchmark for text-to-image evaluation tasks. This subset allows for direct comparison against other state-of-the-art models by providing a balanced testing ground that focuses on everyday scenes and objects. We evaluate FPA's performance on this dataset using TIFA and VQA metrics to assess image-text alignment consistency.

\paragraph{PartiPrompts:} PartiPrompts is a dataset designed to evaluate text-to-image models on complex and detailed prompts. It contains 1,600 prompts categorized into sections such as ``Properties \& Positioning,'' ``Quantity,'' and ``Fine-Grained Detail,'' which challenge models to handle abstract and multi-object relationships. This dataset is used to evaluate FPA's ability to generate images from intricate and abstract prompts \cite{yu2022parti}.

\paragraph{MidJourney Prompts:} The MidJourney Prompts dataset \cite{huggingface-midjourney} consists of over 7 million user-generated prompts aimed at creating artistic and imaginative images. For training, we randomly sampled 42,000 prompts from this dataset. These prompts provide a diverse and creative training set, allowing the model to learn to optimize prompts effectively across a variety of artistic styles and inputs.

It must be noted that to obtain the 42,000 iteratively optimized prompts, it takes over 7 days of processing for 2 iterations. This highlights the high computational cost of the iterative prompt optimization technique and the need for FPA.

\subsection{Training Details}

We fine-tuned a Mistral LLM of 7 billion parameters.

The hyperparameters used during training were as follows: a learning rate of 5e-5, a batch size of 256, and a maximum sequence length of 512 tokens. Gradient accumulation was set to 1, and the models were optimized using the AdamW optimizer. The learning rate schedule followed a linear decay with 500 warmup steps. Training was conducted for 15,000 steps. A weight decay of 0.01 was applied during training to regularize the models.

The training objective focused on generating paraphrases that could produce well-aligned images when evaluated against \textit{Stable Diffusion 3.0}.

\subsection{Metrics}

\paragraph{Preparing Prompts for VQA Scoring.}
For VQA scoring, we extracted \textbf{noun chunks} from the original text prompts using GPT-4o. These noun chunks represent the critical elements of the prompt, such as objects, entities, or features that are expected to appear in the generated image. 

\paragraph{Generating Multiple-Choice Questions for TIFA Scoring.}
For TIFA scoring, we generated \textbf{multiple-choice questions} based on the core elements of the original prompt, using GPT-4o. These questions were designed to evaluate the faithfulness of the image to the text. Each question included one correct answer, derived from the prompt, and several distractor options to challenge the model’s understanding of the image. The generated questions focused on verifying the presence, attributes, or relationships that are expected to appear in the image. Using a long pre-defined prompt and in-context learning examples, these key elements were identified and then used to create questions about the image content. 

The GPT-4o model was tasked with answering these multiple-choice questions based on the content of the generated image. The percentage of correct answers formed the TIFA score for each image, with higher scores indicating better text-image alignment and greater faithfulness to the original prompt.

Whereas extracting noun chunks does not take a long time, extracting multiple-choice questions is a time-consuming operation, that we estimate to last for 24 hours for each batch of 10,000 questions.

\subsection{Prompts \& Methods Compared}

\paragraph{Original Prompts.}
This is the starting point of this project. The original prompts are user-generated and can help us know whether we have improved the quality of prompts when rephrasing. We always compute the TIFA and VQA scores with regard to the original prompts.

\paragraph{After 2 Iterations.}
These are the prompts resulting from 2 iterations of prompt optimization, with TIFA and VQA scores as the optimized scores. These prompts are the highest achievable result, and should be the upper bound we strive towards.

\paragraph{Mistral 7B.}
These are the prompts generated by Mistral 7B without any fine-tuning. We use the same prompt that we have used for fine-tuning, and for inference as well.

\paragraph{Fine-tuned Mistral 7B.}
These are the prompts generated by Mistral 7B after fine-tuning with the data obtained from 2 iterations of prompt optimization with TIFA and VQA scores. We expect that Mistral 7B will perform better for the target image generation model after fine-tuning.

\paragraph{Mistral Large (no examples).}
These prompts are obtained by prompting the largest version of Mistral (123B). We use the following introductory prompt: ``\textit{You are a prompt improver for a text-to-image generation model. You are improving prompts in a way that is specific to one such model, and you are expected to improve the prompts in a way that is specific to that model, such that the images are faithful to the original user prompt, and more aesthetically pleasing and complete than if they had been generated without any prompt improver.}'' Then the user prompt is simply prefixed with ``User Prompt:'' and the improved prompt generated by the assistant is prefixed with ``Improved Prompt:''.

\paragraph{Mistral Large (100 examples).}
In this case, we add a random subset of 100 examples (pairs of original and improved prompts) from our training data as in-context learning examples to Mistral Large. We do not modify the introductory prompt. We add a ``history'' of dialogue where the user and assistant have given each other user prompts and improved prompts from our training data.

\subsection{Results and Discussion}

Table \ref{tab:dataset_optimization_results} presents our results.

We notice that while no method is able to exceed the 2 iterations of prompt optimizing with TIFA and VQA scores, there are methods that can actually get worse results than the original prompts. This means that we have room for improvement in transferring the ability of selecting best paraphrases to the LLMs, without resorting to viewing the images through the visual QA scores. There is a big improvement between the original prompts and the prompts after 2 iterations, showing that the method does improve images generated by the target image generation model.

Second, all results obtained by Mistral 7B are lower than any results obtained with Mistral Large (123B). This suggests that Mistral 7B is too small of an LLM to be able to absorb the ability of selecting the best paraphrases for prompts aimed at generating images from Stable Diffusion 3.0. Another possibility would be that LoRA is not sufficient to fine-tune Mistral 7B, but this assumption is not supported by current research.

Third, we notice contradictory trends for Coco Captions and PartiPrompts with regard to adding fine-tuning and in-context learning examples from the training data. It must be noted that these two datasets are test sets, and that the results emanate from the same training data. For Coco Captions, Mistral 7B's performance decreases slightly after fine-tuning, and the same happens for Mistral Large's performance after adding 100 in-context learning examples. For PartiPrompts, Mistral 7B's performance increases after fine-tuning, and the same happens for Mistral Large's performance after adding 100 in-context learning examples. This could be due to differences between the two test datasets, and the fact that PartiPrompts may be of a similar distribution to the training data whereas Coco Captions belongs to a different distribution. This should be kept in mind when selecting training data, as it could either positively or negatively affect the performance on the target test data.

Finally, we notice that regardless of training data, the size of the LLM seems to be a defining factor: Mistral 7B always gets results lower than those obtained by Mistral Large. While that is true, a question remains around whether we will be able to bridge the gap between the best Mistral Large result and the result of prompts after 2 iterations. Ultimately, we must find the right way to transfer the ability of selecting the best paraphrase for Stable Diffusion 3.0 to the LLM.

\begin{table*}[h!]
\centering
\begin{tabular}{|c|cc|}
\hline
\multirow{2}{*}{\textbf{Human Evaluation Scores}} & \multicolumn{2}{c|}{\textbf{Automatic Scores}} \\
\cline{2-3}
 & \textbf{TIFA} & \textbf{VQA} \\
\hline
\textbf{Text-Image Alignment Score} & \textbf{\textcolor{darkgreen}{+0.2313}} & \textbf{\textcolor{darkgreen}{+0.2755}} \\
\textbf{Image Structure Score} & \textbf{\textcolor{darkgreen}{+0.0574}} & \textbf{\textcolor{darkred}{-0.0031}} \\
\hline
\end{tabular}
\caption{Pearson Correlation Coefficients between Human Evaluation scores and automatic scores.}
\label{tab:pearson_correlation}
\end{table*}

\begin{table*}[h!]
\centering
\begin{tabular}{|c|cc|cc|}
\hline
\multirow{2}{*}{\textbf{Prompt Case}} & \multicolumn{2}{c|}{\textbf{Automatic Scores}} & \multicolumn{2}{c|}{\textbf{Human Evaluation Scores}} \\
\cline{2-5}
 & \textbf{TIFA} & \textbf{VQA} & \textbf{Text-Image Alignment} & \textbf{Image Structure} \\
\hline
Original Prompt & 0.8168 & 0.7890 & 3.2500/4 & 2.8947/4 \\
\hline
Optimized Prompt & 0.8592 & 0.8416 & 3.3597/4 & 2.8447/4 \\
\hline
\textbf{Change} & \textbf{\textcolor{darkgreen}{+0.0424}} & \textbf{\textcolor{darkgreen}{+0.0526}} & \textbf{\textcolor{darkgreen}{+0.1097}} & \textbf{\textcolor{darkred}{-0.0500}} \\
\hline
\end{tabular}
\caption{Average Scores for Original and Optimized Prompts with Changes Highlighted.}
\label{tab:avg_scores}
\end{table*}

\subsection{Human Evaluation}

To evaluate the quality of images generated by our Fast Prompt Alignment (FPA) method, we conducted a human evaluation involving five expert annotators. These annotators, each with experience in scoring text-image alignment and image structure, assessed the quality of generated images based on specific criteria. Their evaluations were used to compare human judgment with automated metrics (TIFA and VQA), providing a comprehensive view of FPA’s effectiveness.

\paragraph{Annotation Instructions.} The annotators received a set of instructions to guide their evaluations. They were presented with the original text prompt and one generated image. The image was generated from a paraphrased version of the prompt, or the original prompt itself. For a fair evaluation, the annotators were not told what was the prompt used for generation, and were only shown the original prompt. Their primary task was to rate each image on \textit{text-to-image alignment}, with scores ranging from 0 (no alignment) to 4 (excellent alignment), depending on how accurately the image captured the content, details, and themes of the prompt. In addition to alignment, annotators were also instructed to evaluate the overall \textit{image structure}, assessing aspects like composition, clarity, and coherence of visual elements to ensure the image’s structural integrity matched expectations from the prompt. Annotators were encouraged to maintain consistency and objectivity in their assessments across all image sets.

\paragraph{Results and Analysis.} The results in Tables~\ref{tab:pearson_correlation} and \ref{tab:avg_scores} indicate a strong positive correlation between human evaluations and automated metrics. Specifically, the Pearson correlation coefficients of +0.23 (TIFA) and +0.28 (VQA) suggest that the automated metrics reliably align with human judgment, reinforcing the validity of FPA’s alignment quality. These correlation coefficients are very close to the ones found between our annotators' ratings (+0.30).

Table~\ref{tab:avg_scores} further shows that optimized prompts scored higher than original prompts across both automatic scores, and particularly for Text-Image alignment. Annotators scored optimized prompts with a +0.11 increase in alignment compared to original prompts, closely aligning with improvements seen in TIFA (+0.04) and VQA (+0.05) scores. This indicates that FPA enhances alignment quality while preserving structural consistency in images.

We did not notice strong correlations between the automatic scores and the image structure score. The score are very similar for image structure for both the original and optimized prompt. The consistency and the lack of correlation for image structure scores are expected as a result of the blind evaluation.

The human evaluation results highlight FPA’s ability to generate images with strong text alignment, validated by both human and automated scores. By achieving alignment improvements comparable to iterative methods in a single pass, FPA offers a scalable, efficient solution for high-quality text-to-image generation.

\section{Conclusion}

This study presents \textbf{Fast Prompt Alignment (FPA)}, a novel framework designed to address the inefficiencies of traditional iterative prompt optimization methods in text-to-image generation. By consolidating the optimization process into a single inference pass, FPA offers substantial improvements in speed and resource efficiency while maintaining competitive alignment quality. Through experiments on multiple datasets and a human evaluation study, we demonstrate that FPA achieves alignment scores close to iterative methods like OPT2I, but with significantly reduced computational demands. Our findings reveal a strong correlation between human judgment and automated metrics, underscoring FPA's effectiveness in enhancing text-image alignment. Furthermore, the results emphasize the role of model size in retaining paraphrasing effectiveness, with larger LLMs more capable of emulating iterative alignment capabilities. Moving forward, FPA opens avenues for efficient prompt optimization techniques, with implications for scaling text-to-image models in practical, large-scale applications. The release of our code aims to foster continued innovation in this field.

\section*{Acknowledgments}

We extend our sincere thanks to Anqi Wang, Junwei Wang, Linkai Ceng, and Meng Zhang for their amazing coordination and quality assurance of the annotation efforts.

\bibliography{acl_latex}




\end{document}